# The CriticalSet problem: Identifying Critical Contributors in Bipartite Dependency Networks

Sebastiano A. Piccolo[1][0000−0002−6986−3344]
Andrea Tagarelli[2][0000−0002−8142−503X]

[1] University of Calabria, DeMaCS – Dept. Mathematics and Computer Science
[2] University of Calabria, DIMES – Dept. Computer Engineering, Modeling, Electronics, and Systems Engineering
{sebastiano.piccolo, andrea.tagarelli}@unical.it

**Abstract.** Identifying critical nodes in complex networks is a fundamental task in graph mining. Yet, methods addressing an all-or-nothing *coverage* mechanics in a bipartite dependency network— a graph with two types of nodes where edges represent dependency relationships across the two groups only—remain largely unexplored. We formalize the CriticalSet problem: given an arbitrary bipartite graph modeling dependencies of items on contributors, identify the set of $k$ contributors whose removal isolates the largest number of items. We prove that this problem is NP-hard and requires maximizing a supermodular set function, for which standard forward greedy algorithms provide no approximation guarantees. Consequently, we model CriticalSet as a coalitional game, deriving a closed-form centrality, `ShapleyCov`, based on the Shapley value. This measure can be interpreted as the expected number of items isolated by a contributor's departure. Leveraging these insights, we propose `MinCov`, a linear-time iterative peeling algorithm that explicitly accounts for connection redundancy, prioritizing contributors who uniquely support many items. Extensive experiments on synthetic and large-scale real datasets, including a Wikipedia graph with over 250 million edges, reveal that `MinCov` and `ShapleyCov` significantly outperform traditional baselines. Notably, `MinCov` achieves near-optimal performance, within 0.02 AUC of a Stochastic Hill Climbing metaheuristic, while remaining several orders of magnitude faster.

## 1 Introduction

Many real-world systems and online platforms including Wikipedia, open-source software repositories, review websites, and collaborative communities, rely critically on the contributions of their participants. In practice, however, contributions are highly uneven: a small fraction of users typically produces a disproportionate share of the content. This concentration of activity has important consequences for both the robustness and the reliability of these systems. When a platform depends heavily on a limited set of contributors, it becomes more vulnerable to failures, strategic behavior, or systematic biases in the information provided. Consequently, identifying the key contributors and quantifying how strongly a system depends on them are fundamental steps toward understanding influence, responsibility, and information concentration.

We model these systems as *bipartite dependency networks*, where one node set, the contributors, is connected to the other node set, the items, where the latter have a functional dependency on the contributors. Typical examples of such network span different domains, such as code dependencies of software applications on libraries, supply dependencies of industries on suppliers, or purchase dependencies of products on customers.

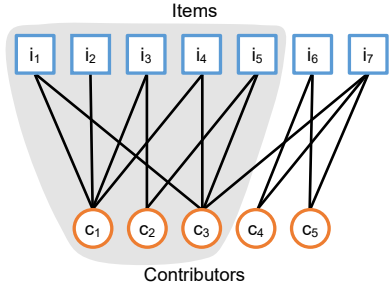

**Fig. 1.** An example of bipartite dependency graph of contributors ($c_i$) and items ($i_j$). The shaded subgraph denotes to the Critical Set of size 3, which corresponds to the set $\{c_1, c_2, c_3\}$, as their removal causes the isolation of items $i_1$–$i_5$.

To identify the contributors on whom many items critically depend, we formalize the CRITICALSET problem. Given a bipartite graph and a budget $k$, the objective is to select $k$ contributors that maximize the number of items whose contributors are *all* selected (Figure 1). In this way, CRITICALSET provides a direct approach to unveil the critical dependencies of items on specific contributors.

We argue that standard network-analysis tools do not adequately capture this notion of functional dependency. Path-based centrality measures, such as betweenness, PageRank or eigenvector centrality, emphasize global connectivity and diffusion, which are often less meaningful in settings dominated by localized contributor-item interactions. Conversely, purely local measures such as node degree ignore redundancy: contributing to an item already supported by many others is less critical than being its sole or one of few contributors. Similarly, classic robustness metrics and percolation approaches that focus on connectivity or network fragmentation fail to capture the relevant notion of vulnerability here, which is determined by item coverage rather than structural disconnection.

From an algorithmic perspective, CRITICALSET is challenging. We show that the problem is NP-hard and inherits the strong inapproximability of the Densest $k$-Subgraph problem, ruling out efficient exact or constant-factor approximation algorithms under standard assumptions. Moreover, the objective is supermodular, and therefore the classic greedy approximation guarantees that rely on submodularity in influence maximization and coverage problems do not apply.

To address these challenges, we adopt a game-theoretic and algorithmic viewpoint centered on contributors' *pivotality*. Essentially, a contributor is pivotal if they represent the final point of failure needed to completely isolate an item (Figure 1). We exploit this structural property in two complementary ways. First, we model CRITICALSET as a coalitional game and derive an exact closed-form expression for the Shapley value, yielding a principled centrality that quantifies each contributor's expected pivotality. Second, we design `MinCov`, a linear-time iterative peeling algorithm that deterministically prioritizes contributors with low marginal impact, producing high-quality approximations in practice.

Beyond contributor–item systems, in our experiments we applied our framework to web trackers and word-document networks as well. Our experiments show that our Shapley value centrality and `MinCov` consistently yield critical sets of contributors that cover a larger number of items compared to baselines including forward-greedy maximization, PageRank, betweenness, and $k$-core decomposition. Our findings confirm that diverse real-world systems depend disproportionately on a small fraction of critical contributors, highlighting a significant structural vulnerability. Identifying such critical sets provides a direct measure of dependence, vulnerability, and concentration of influence.

**Contributions.** Our main contributions are:

- We introduce the CRITICALSET problem, a new all-or-nothing coverage formulation designed to identify functionally critical nodes in bipartite networks.
- We prove NP-hardness and approximation hardness via a reduction from Densest $k$-Subgraph, and characterize structural properties of the objective.
- We model the problem as a coalitional game and derive an exact closed-form Shapley-value centrality computable in $O(|E|)$ time.
- We design `MinCov`, a linear-time iterative peeling algorithm that generalizes $k$-core decomposition and scales to large graphs.
- We conduct an extensive empirical evaluation on diverse synthetic and real-world dependency networks. Our results demonstrate that our algorithms consistently outperform traditional baselines and achieve near-optimal performance: reaching within 0.02 AUC of a Stochastic Hill Climbing metaheuristic while being up to three orders of magnitude faster.

Together, these results provide both theoretical foundations and scalable tools for analyzing criticality and dependence in large bipartite networks. *We are committed to reproducible research and release our code and evaluation data upon acceptance of this work.*

## 2  Related Work

The CRITICALSET problem is conceptually related to node ranking and top-$k$ selection in networks, including centrality-based rankings [23, 22], influence maximization, and coverage problems [12, 5, 6, 17]. However, unlike influence maximization under the Independent Cascade or Linear Threshold diffusion models, or maximum coverage, where an item is considered affected or activated if at least one selected node reaches it through the diffusion process, CRITICALSET requires all contributors of an item to be selected. As we show in Section 4, this difference makes the objective of CRITICALSET supermodular, in contrast to the submodular objective of influence maximization.

From an optimization perspective, CRITICALSET generalizes densest-$k$-subgraph [10, 11, 20] and node-removal problems on bipartite graphs, including variants of network resilience formulations and percolation approaches [1, 2]. However, these problems typically aim to maximize internal density or to fragment connectivity, whereas our objective focuses on preserving item functionality under contributor selection, which is orthogonal to graph fragmentation and leads to different optimal solutions.

The problem is also related to the computation of project robustness (e.g. the bus factor in software engineering) and redundancy analysis in collaborative systems [4, 28, 15, 27], where degree-based heuristics are commonly used to identify critical contributors. Our theoretical and empirical results show that such heuristics can be highly suboptimal, motivating the need for principled formulations and algorithms that explicitly account for contributor redundancy and full item coverage.

Finally, unlike common projection-based approaches that collapse bipartite structures into cliques [26], thereby losing information and creating spurious links [8], our method performs a direct analysis that preserves the full dependency mechanics of the original network.

## 3   The Critical Set Problem

**Preliminaries.** Let $G = (V, E)$ be an undirected graph, with $V$ being the set of nodes and $E$ the set of edges. For a node $v$, we denote its neighborhood (i.e. the set of nodes connected to $v$) by $\Gamma(v)$ and its degree by $\deg(v) = |\Gamma(v)|$. Given a graph $G = (V, E)$, for $S \subseteq V$ the induced subgraph $G[S]$ contains the nodes in $S$ and all the edges with both endpoints in $S$ (i.e., the set of edges of $G[S]$ is $E(G[S]) = \{(u, v) \in E \mid u \in S \land v \in S\}$). A graph is *bipartite* if its node set can be partitioned into two disjoint sets $V_1$ and $V_2$, such that $V = V_1 \cup V_2$, $V_1 \cap V_2 = \emptyset$, and every edge connects a node in $V_1$ to one in $V_2$.

**The CriticalSet Problem.** We consider a specific class of heterogeneous networks consisting of two types of nodes: contributors (e.g., individuals) and items (e.g., products, opinions). Since our focus is exclusively on contributor–item relationships, we model these structures as bipartite graphs $B = (C, I, E)$ where $C$ denotes contributors and $I$ items, and $E \subseteq C \times I$. Given $S \subseteq C$, an item $i \in I$ is fully covered if $\Gamma(i) \subseteq S$. That is, if $S$ contains all neighbors of $i$. Finally, let $\mathrm{cov}(S) = |\{i \in I : \Gamma(i) \subseteq S\}|$ denote the number of items fully covered by $S$ (i.e., the number of items that would become isolated if the contributors in $S$ were removed from $B$).

**Definition 1 (The CriticalSet problem).** *Given a bipartite graph $B = (C, I, E)$ and a positive integer $k$, the CriticalSet problem asks for a set $S^* \subseteq C$ of size $|S^*| \leq k$ that maximizes $\mathrm{cov}(S^*)$. Formally:*

$$S^* = \arg \max_{S \subseteq C, |S| \leq k} \mathrm{cov}(S)$$

*Any maximizing set is called a critical set.*

## 4   Hardness of the CriticalSet Problem

We establish the computational hardness of CriticalSet via a reduction from the Densest $k$-Subgraph (DkS) problem, which is known to be NP-hard [10]. Given a simple graph $G = (V, E)$ and an integer $k$, DkS asks for a subset $X \subseteq V$ with $|X| \leq k$ maximizing the number of induced edges $|E(G[X])|$.

The reduction is intuitive: each vertex of $G$ becomes a contributor and each edge becomes an item; covering an item then requires selecting both endpoints of the corresponding edge.

**Theorem 1.** *CriticalSet is NP-hard and at least as hard to approximate as DkS. In particular, any approximation guarantee for one problem transfers to the other with the same factor.*

*Proof.* Let $x = (G = (V, E), k)$ be an instance of DkS. We construct a bipartite graph $B = (C, I, E')$ as follows: $C = V$, $I = E$, and

$$E' = \{(u, e) \mid u \text{ is an endpoint of edge } e\}.$$

Thus each item is adjacent exactly to the two contributors corresponding to the endpoints of the original edge. The construction clearly takes $O(|V| + |E|)$ time.

Consider any set $S \subseteq C$ with $|S| \leq k$. An item corresponding to edge $(u, v) \in E$ is fully covered if and only if both $u$ and $v$ belong to $S$. Therefore,

$$\mathrm{cov}(S) = |E(G[S])|.$$

Hence maximizing coverage in $B$ is equivalent to maximizing the number of induced edges in $G$, and optimal solutions have identical objective values. This establishes an approximation-preserving reduction [9] (specifically an S-reduction) from DkS to CriticalSet, implying both NP-hardness and the transfer of inapproximability results. □

The reduction extends naturally to multigraphs by treating parallel edges as distinct items, and to hypergraphs via their standard bipartite incidence representation. The reduction also transfers the strong inapproximability of DkS. Despite extensive study, DkS admits neither a PTAS nor any known constant-factor approximation, with the best algorithms achieving only $O(n^{1/4+\epsilon})$ guarantees [11, 7, 18]. Stronger lower bounds hold under ETH and the Small Set Expansion hypothesis [24, 29]. Consequently, CriticalSet inherits the same hardness of approximation.

*Supermodularity of CriticalSet.* We also provide a further insight into the hardness of CriticalSet, showing that the function $\text{cov}(\cdot)$ is supermodular.

**Lemma 1.** *The function* $\text{cov} : 2^C \to \mathbb{N}$ *is monotone and supermodular.*

*Proof.* For each item $i \in I$, let $f_i(S) = \mathbb{I}(\Gamma(i) \subseteq S)$. Then $\text{cov}(S) = \sum_{i \in I} f_i(S)$.

Fix an item with neighborhood $T = \Gamma(i)$. For any $A \subseteq B \subseteq C$ and $v \notin B$, adding $v$ can only complete the containment $T \subseteq S$ if all other elements of $T$ are already present. Hence

$$f_i(A \cup \{v\}) - f_i(A) \leq f_i(B \cup \{v\}) - f_i(B),$$

so $f_i$ is supermodular. Since sums of supermodular functions are supermodular, cov is supermodular. Monotonicity is immediate. □

This contrasts with submodular objectives that admit strong greedy approximation guarantees. The supermodular (increasing-returns) structure of CriticalSet aligns instead with clique-like objectives and helps explain its strong hardness of approximation.

## 5 Approximating CriticalSet

By studying the CriticalSet problem as a coalitional game, we uncover a key structural property: coverage increases only when the last missing contributor of an item is selected, i.e., value is created at pivotal events. This naturally suggests prioritizing contributors that are often pivotal.

We formalize this idea in two ways. First, we derive a closed form solution for the Shapley value of the CriticalSet game, obtaining a centrality score, `ShapleyCov`, that quantifies expected pivotality of contributors over uniformly random orderings. Second, we develop an algorithm, `MinCov`, that implements the same principle deterministically through iterative peeling, repeatedly removing the contributor with the smallest marginal impact on coverage.

### 5.1 CriticalSet as a coalitional game

We model contributors as players in a coalitional game where the value of a coalition equals the number of items it fully covers. This perspective assigns each contributor a score proportional to its marginal contribution to coverage.

A coalitional game is specified by a characteristic function $v : 2^C \to \mathbb{R}$ with $v(\emptyset) = 0$. We adopt the *Shapley value*, which assigns to each player its average marginal contribution over all arrival orders:

$$\phi_i = \frac{1}{|C|!} \sum_{\pi \in \Pi_C} \big(v(B(i,\pi) \cup \{i\}) - v(B(i,\pi))\big). \tag{1}$$

where $\Pi_N$ denotes the set of all permutations of $C$, and $B(i,\pi)$ is the set of contributors that precede $i$ in the permutation $\pi$.

The Shapley value is uniquely characterized by standard fairness axioms and therefore provides a canonical (rather than heuristic) allocation of utility. Although computing it is generally #P-hard, the structure of CRITICALSET yields an exact closed-form solution. We define the game by

$$v(S) = \text{cov}(S) = \sum_{i \in I} \mathbb{I}(\Gamma(i) \subseteq S), \quad S \subseteq C. \tag{2}$$

Here players are contributors, while items only determine coalition utility; maximizing $v(S)$ under $|S| \le k$ recovers CRITICALSET, and $v(C) = |I|$.

**Theorem 2 (Shapley value for CRITICALSET).** *The Shapley value of the coalitional game $(C, v)$ admits the following closed form*

$$\phi_c = \sum_{i \in \Gamma(c)} \frac{1}{\deg(i)}, \quad \forall c \in C. \tag{3}$$

*Proof.* Since the Shapley value is linear and the characteristic function is

$$v(S) = \sum_{i \in I} v_i(S) \quad \text{with } v_i(S) = \mathbb{I}(\Gamma(i) \subseteq S),$$

it suffices to analyze each $v_i(\cdot)$. Contributor $c \in \Gamma(i)$ has a marginal contribution of one if and only if it is the last neighbor of $i$ in a uniformly random permutation, which occurs with probability $1/\deg(i)$. Summing these probabilities over all items incident to $c$ yields equation (3). □

The Shapley value in equation (3) is effectively a centrality measure, which we call `ShapleyCov`, that can be computed in $O(|E|)$ time in a single pass over the edges and is trivially parallelizable. It can be interpreted as a refined degree measure: importance increases with the number of supported items but decreases with redundancy, prioritizing contributors that are critical for coverage.

### 5.2 An iterative peeling procedure to approximate CRITICALSET

The pivotality interpretation and the supermodularity of the objective suggest a reverse-greedy strategy: when seeking any size-$k$ Critical Set, *contributors that are connected to few items are unlikely to be pivotal and should be removed first*. This leads to an iterative peeling procedure (Algorithm 1) that repeatedly removes the contributor covering the fewest items (lines 4–6) and updates remaining counts (lines 7–12). The final ordering $\pi$ is obtained by reversing this removal sequence (line 13), such that the first $k$ contributors represent the most critical nodes for a given budget. By using a bucket queue (line 2), a specialized priority queue where an array $Q$ stores nodes in lists indexed by their current coverage, `MinCov` maintains $O(1)$ operations for insertion, extraction, and priority updates, achieving an $O(|E|)$-time implementation. Conceptually, `MinCov` is a greedy analogue of the `ShapleyCov` ordering: both prioritize contributors according to their likelihood of being pivotal, the former adaptively and the latter in expectation.

**Algorithm 1:** Minimum Coverage (`MinCov`)

**Input:** a bipartite graph $B = (C, I, E)$
**Output:** Ordering $\pi$ of contributors

1  `inserted[c]` $\leftarrow 0$, $\forall c \in C$;   `covered[i]` $\leftarrow 0$, $\forall i \in I$;   $\pi \leftarrow []$
2  $Q \leftarrow$ BucketQueue($B$)                                                 // Prioritizes contributors by lowest item coverage
3  **while** $Q$ *not empty* **do**
4  $\quad c \leftarrow Q.\text{pop}()$                                              // Gets the contributor with the lowest coverage
5  $\quad \pi.\text{append}(c)$                                                     // Appends $c$ to the ordering $\pi$
6  $\quad$ `inserted[c]` $\leftarrow 1$                                             // Marks the contributor $c$ as processed
7  $\quad$ **foreach** $i \in B.neighbors(c)$ **do**
8  $\quad\quad$ **if** *not* `covered[i]` **then**                                  // Decreases by one the coverage of all
9  $\quad\quad\quad$ **foreach** $c' \in B.neighbors(i)$ **do**                    // contributors $c'$ connected to
10 $\quad\quad\quad\quad$ **if** *not* `inserted[c']` **then**                      // the items now covered by $c$
11 $\quad\quad\quad\quad\quad$ $Q.\text{decrease\_coverage}(c')$
12 $\quad\quad$ `covered[i]` $\leftarrow 1$                                         // Marks all items connected to $c$ as fully covered

13 **return** $\pi.\text{reverse}()$

**Relations between `MinCov` and $k$-core decomposition.** `MinCov` can be viewed as a strict generalization of the classical minimum-degree peeling procedure underlying the $k$-core decomposition. Applying the standard incidence transformation that maps edges to items reduces $k$-core exactly to `MinCov`. However, coverage differs fundamentally from degree: an item is removed after being covered once, which *discounts redundant connections* and better captures criticality. Consequently, while `MinCov` can recover $k$-cores as a special case, the reverse is not true, and the resulting rankings differ substantially in practice, as highlighted by our subsequent experiments.

## 6 Experimental setup

**Real-world networks.** To assess the performance of our CriticalSet methods and competing ones, we employ the following twelve large-scale bipartite networks obtained from: ($i$) the actor-movie network dbpedia [3], ($ii$) the GitHub 2009 competition dataset [30] connecting developers to repositories, ($iii$) the bag-of-words of NeurIPS papers [21] connecting words to papers, ($iv$) the bag-of-word of Wikipedia's excellent articles [19] connecting words to articles, ($v$) the user ratings of stories on the social media platform Digg [14] connecting users to stories, ($vi$) Amazon user–item ratings [16], ($vii$) the affiliations of users to groups on Flickr [25], ($viii$) the MovieLens 10M dataset [13] connecting users to movies via their ratings, ($ix$) the Open Academic Graph (OAG) [32] connecting researchers to papers, ($x$) the affiliation of users to groups on LiveJournal [25], ($xi$) web domains and the trackers they contain [31], and ($xii$) the Wikipedia edit history (2001–2017) [19].

Table 1 summarizes the structural properties of our real-world networks, including power-law exponents ($\alpha$), degree-1 fractions ($\phi$), and the ratio of *unique* contributors ($\gamma_C$) who provide sole coverage for at least one item. These metrics capture the inherent difficulty of the CriticalSet instance. For example, `github` exhibits a "shallow" structure where 69% of items possess only a single contributor ($\phi_I$), making it a trivial target for greedy heuristics. In contrast, datasets like `bag-nips` exhibit high redundancy of contributors (i.e., $\phi_I = 0$), thus forming a complex *supermodular core*—that is, a subgraph where the fraction of degree-one items is close to zero. This way, critical sets become apparently "functionally invisible" to greedy approaches.

Table 1. Statistics of real bipartite datasets.

| Dataset | $|C|$ | $|I|$ | $|E|$ | $\bar{k}_C$ | $\bar{k}_I$ | $\alpha_C$ | $\alpha_I$ | $\phi_C$ | $\phi_I$ | $\gamma_C$ |
|---|---|---|---|---|---|---|---|---|---|---|
| dbpedia | 81 085 | 76 099 | 281 396 | 3.47 | 3.70 | 0.19 | 0.32 | 0.59 | 0.20 | 0.12 |
| github | 56 519 | 120 867 | 440 237 | 7.79 | 3.64 | 0.17 | 0.13 | 0.41 | 0.69 | 0.64 |
| bag-nips | 12 375 | 1 500 | 746 316 | 60.31 | 497.54 | 0.23 | 1.54 | 0.03 | 0.00 | 0.00 |
| excellent | 273 959 | 2 780 | 2 941 902 | 10.74 | 1058.24 | 0.14 | 0.79 | 0.69 | 0.00 | 0.00 |
| digg-votes | 139 409 | 3 553 | 3 018 197 | 21.65 | 849.48 | 0.13 | 0.27 | 0.28 | 0.00 | 0.00 |
| amazon | 2 146 057 | 1 230 915 | 5 838 041 | 2.72 | 4.74 | 0.11 | 0.14 | 0.69 | 0.50 | 0.16 |
| flickr | 395 979 | 103 631 | 8 545 307 | 21.58 | 82.46 | 0.16 | 0.12 | 0.34 | 0.24 | 0.05 |
| movielens | 69 878 | 10 677 | 10 000 054 | 143.11 | 936.60 | 0.22 | 0.18 | 0.00 | 0.01 | 0.00 |
| oag | 9 381 152 | 6 736 186 | 22 766 556 | 2.43 | 3.38 | 0.14 | 0.14 | 0.60 | 0.30 | 0.16 |
| livejournal | 3 201 203 | 7 489 073 | 112 307 385 | 35.08 | 15.00 | 0.34 | 0.07 | 0.06 | 0.71 | 0.42 |
| trackers | 12 756 244 | 27 665 730 | 140 613 762 | 11.02 | 5.08 | 0.06 | 0.08 | 0.61 | 0.31 | 0.05 |
| wikipedia | 8 116 897 | 42 640 545 | 255 709 660 | 31.50 | 6.00 | 0.07 | 0.09 | 0.59 | 0.47 | 0.19 |

**Synthetic Network Generation.** To assess proximity to optimality, we generated synthetic bipartite graphs using the bipartite configuration model. We consider six configurations targeting different structural regimes:

- Power-law graphs (a–e): We vary the exponents ($\alpha$) and maximum degrees ($D$) to control heterogeneity. Configurations (a–b) represent asymmetric item/contributor maximum degrees; (c) is balanced; (d–e) vary the skeweness of the degree distributions of the node sets.
- Random graph (f): An Erdős-Rényi (ER) bipartite graph ($p = 0.004$) providing a non-structured baseline.

For the power-law instances, lower $\alpha$ values correspond to higher heterogeneity, while $D$ constrains the hub sizes. For all networks, we set $|C| = |I| = 5\,000$; all generation parameters are reported in Table 3.

**Competing methods.** We compare `MinCov` and `ShapleyCov` against the following baselines, each of which was computed on the set of contributors $C$ of an inout graph: degree centrality (**DC**), betweenness centrality (**BC**), PageRank (**PR**), the minimum degree peeling procedure used to compute the $k$-core decomposition and to approximate the densest subgraph (**DS**) applied directly to the bipartite graph, and a forward greedy algorithm that maximizes $cov(\cdot)$ (**FG**). We also considered closeness and eigenvector centrality, but they exhibited poor results. Finally, due to the scale of the real-world datasets, we adopt a Stochastic Hill Climbing (**SHC**) metaheuristic exclusively for synthetic graphs. This serves as a reference to estimate the heuristics' proximity to optimality and compare their running times.

**Assessment criteria.** Recall that a critical set is a subset of contributors of size $k$ that covers the largest number of items or, equivalently, a subset of contributors of size $k$ whose removal maximally reduces the number of covered items.

To evaluate our CriticalSet methods and competing ones, we measure the cumulative number of items in $I$ covered by the selected contributors as the size $k$ of the critical set increases—this produces a coverage curve as a function of $k$. More specifically, we measure the normalized area under such a coverage curve (AUC). Given a bipartite graph $B = (C, I, E)$ and a ranking $R$ of contributors in $C$ produced by a method for solving the CRITICALSET task, the area under the coverage curve is defined as: $\text{AUC}(B, R) = \frac{1}{|C|} \sum_{i=1}^{|C|} \frac{\text{cov}(B, R_{\leq i})}{|I|}$, where $R_{\leq i}$ contains the first $i$ contributors in $R$ and $\text{cov}(B, R_{\leq i})$ yields the number of items fully covered by $R_{\leq i}$. $\text{AUC}(B, R)$ ranges between 0 and 1, with higher values corresponding to better performance in approximating critical sets.

**Table 2.** Area under the coverage curve for critical sets of real-world graphs (higher values correspond to better solutions). Best in bold; second best underlined.

| Dataset | PageRank | Degree | Fwd. Greedy | Dens. Subgraph | ShapleyCov | MinCov |
|---|---|---|---|---|---|---|
| dbpedia | 0.661 | 0.675 | **0.722** | 0.636 | <u>0.709</u> | 0.699 |
| github | 0.736 | 0.724 | **0.757** | 0.652 | <u>0.751</u> | **0.757** |
| bag-nips | 0.073 | 0.078 | 0.002 | 0.079 | <u>0.080</u> | **0.148** |
| excellent | 0.204 | 0.026 | 0.021 | 0.037 | <u>0.492</u> | **0.507** |
| digg-votes | 0.074 | <u>0.155</u> | 0.001 | <u>0.155</u> | **0.348** | **0.348** |
| amazon | 0.745 | 0.647 | 0.758 | 0.550 | **0.813** | <u>0.794</u> |
| flickr | 0.525 | 0.440 | 0.448 | 0.402 | <u>0.658</u> | **0.659** |
| movielens | 0.241 | 0.242 | 0.216 | 0.232 | <u>0.482</u> | **0.618** |
| oag | 0.576 | 0.551 | <u>0.624</u> | 0.465 | **0.625** | 0.617 |
| livejournal | 0.788 | 0.719 | 0.825 | 0.636 | <u>0.833</u> | **0.837** |
| trackers | 0.939 | 0.935 | 0.956 | 0.914 | **0.970** | <u>0.969</u> |
| wikipedia | 0.955 | 0.950 | **0.964** | 0.934 | <u>0.963</u> | <u>0.963</u> |

In addition, we plot a representative selection of curves that will provide punctual information for each value of the budget $k$, complementing the aggregate information of the AUC.

## 7 Results

**Evaluation on the real-world networks.** Here, we assess the performance of MinCov and ShapleyCov contrasting them with the other baselines. Due to its computational complexity, Betweenness Centrality (BC) is computationally prohibitive for our largest real-world datasets. However, as demonstrated in the following sections on synthetic networks, BC consistently underperforms in the CRITICALSET task, justifying its omission from the large-scale evaluation. We report the AUC values in Table 2.

Overall, MinCov and ShapleyCov emerge as the most robust methods, achieving AUC values significantly higher than the competing methods. MinCov attains the highest AUC on seven datasets and ranks second on three, while ShapleyCov performs best on four and ranks second on all others. Conversely, Forward Greedy (FG) is highly inconsistent: while competitive on shallow graphs with high fractions of degree-one items (cf. Table 2, $\phi_I$) and contributors ($\phi_C$), such as dbpedia and github, it exhibits the worst performance on four high-redundancy datasets (cf. Sect. 6), including movielens and excellent. On digg-votes, FG fails to detect any critical set within the budget (Fig. 2), confirming its ineffectiveness on complex supermodular problems.

The transition between the two regimes is visible in the amazon graph (Fig. 2): FG remains competitive on small $k$, as long as it captures *low-hanging* fruits ($\phi_I = 0.5$). As soon as FG hits the *supermodular core* of items with high redundancy of contributors its performance degrades. In contrast, MinCov and ShapleyCov maintain strong performance across all graphs, and emerge as clear winners on graphs with absence of degree-one items ($\phi_I \approx 0$), where FG fails to detect meaningful critical sets (e.g.,

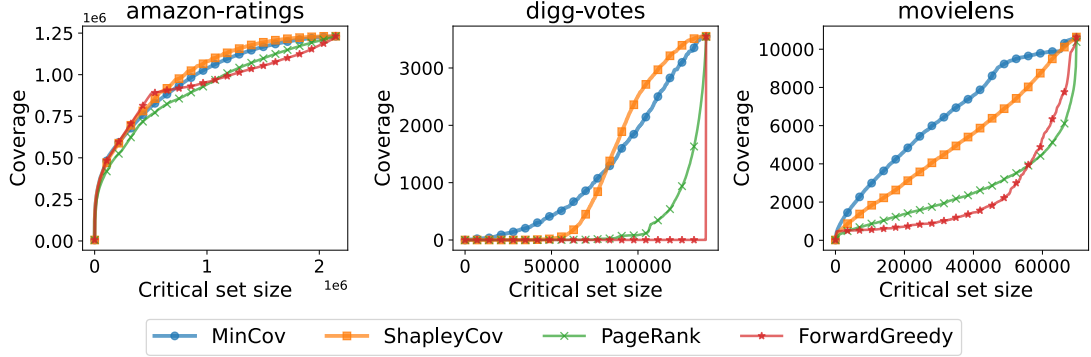

**Fig. 2.** Coverage curves on real datasets (selection).

**Table 3.** AUC for CRITICALSET approximation approaches on synthetic graphs. In bold, the best performer; underlined, the second best. SHC (blue) as reference.

| Configuration | PR | BC | DC | FG | DS | ShapleyCov | MinCov | SHC |
|---|---|---|---|---|---|---|---|---|
| (**a**) $\alpha_C = 0.5$, $\alpha_I = 0.5$, $D_C = 20$, $D_I = 100$ | 0.552 | 0.564 | 0.531 | 0.574 | 0.524 | <u>0.600</u> | **0.601** | 0.615 |
| (**b**) $\alpha_C = 0.5$, $\alpha_I = 0.5$, $D_C = 100$, $D_I = 20$ | 0.560 | 0.547 | 0.547 | 0.536 | 0.545 | <u>0.584</u> | **0.593** | 0.611 |
| (**c**) $\alpha_C = 0.5$, $\alpha_I = 0.5$, $D_C = D_I = 100$ | 0.266 | 0.265 | 0.263 | 0.182 | 0.262 | <u>0.279</u> | **0.311** | 0.323 |
| (**d**) $\alpha_C = 0.5$, $\alpha_I = 0.7$, $D_C = D_I = 100$ | 0.314 | 0.311 | 0.311 | 0.203 | 0.309 | <u>0.327</u> | **0.353** | 0.366 |
| (**e**) $\alpha_C = 0.7$, $\alpha_I = 0.5$, $D_C = D_I = 100$ | 0.314 | 0.316 | 0.308 | 0.254 | 0.307 | <u>0.336</u> | **0.355** | 0.368 |
| (**f**) ER graph ($p = 0.004$) | 0.078 | 0.080 | 0.077 | 0.078 | 0.076 | <u>0.081</u> | **0.133** | 0.133 |

`digg-votes` and `movielens`, Fig 2). The low AUC of all methods on the `bag-nips` dataset arises because items (documents) have high degrees (many words), making full coverage difficult, particularly with a small budget $k$. These results indicate that structural properties beyond degree distributions, specifically redundancy depth ($\phi_I$, $\phi_C$, $\gamma_C$), dictate the relative difficulty of CRITICALSET instances, pointing to an interesting direction for future investigations.

A one-sided paired Wilcoxon signed rank test indicates that `MinCov` and `ShapleyCov` achieve a significantly larger AUC than FG and the other baselines ($p < 0.05$). The test between `MinCov` and `ShapleyCov` does not detect any statistically significant difference in AUC ($p = 0.52$).

**Optimality and approximation gap.** We evaluate our methods against a Hill-Climbing metaheuristic (SHC) run until convergence to estimate the proximity to optimality. Table 3 reports the AUC values, while Fig. 3 provides a zoomed-in comparison of the coverage curves.

`MinCov` and `ShapleyCov` are consistently the best-performing methods, with `MinCov` achieving near-optimal results, trailing SHC by less than 0.02 AUC points across all settings. Notably, `ShapleyCov` remains highly competitive (within 0.05 of SHC), which is remarkable given it is a single-pass centrality measure. In contrast, all other baselines exhibit significantly larger approximation gaps.

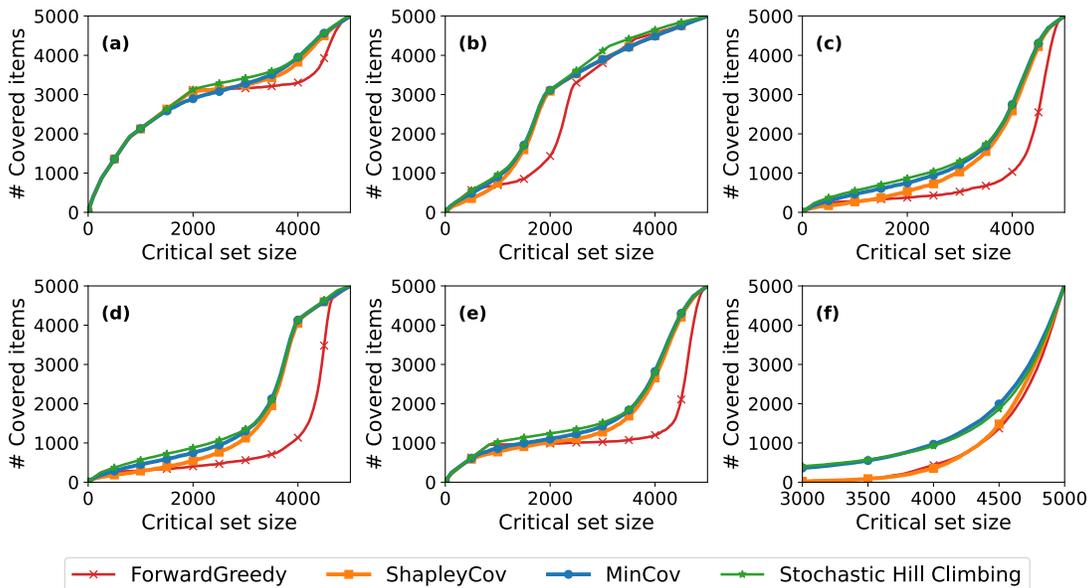

**Fig. 3.** Zoomed-in comparison between `MinCov`, `ShapleyCov`, Forward Greedy and SHC. The letter indicates the synthetic graph configuration (refer to Table 3).

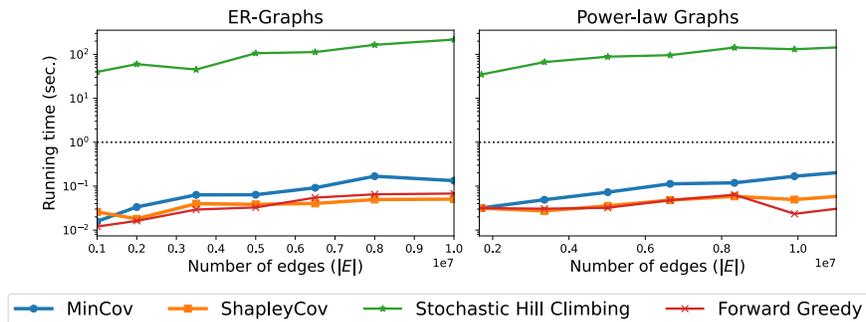

**Fig. 4.** Running time comparison between `MinCov`, `ShapleyCov`, FG, and SHC.

These results also highlight the inadequacy of standard centralities for CRITICALSET: Forward Greedy (FG) is consistently outperformed, and Betweenness Centrality (BC) offers no advantage over simpler degree-based metrics. Finally, the striking performance gap between `MinCov` and Densest Subgraph (DS) confirms that while both utilize iterative peeling, the CRITICALSET objective requires the specific redundancy-aware logic of our approach.

**Scalability and price of optimality.** We have seen that `MinCov` and `ShapleyCov` obtain excellent results. In particular, `MinCov` is near optimal w.r.t. SHC. Here, we perform a scalability analysis to evaluate the *price of optimality* in terms of computation time.[3] To this end, we generate random bipartite graphs (ER and Power-law) with 50 000 nodes per node set and a total number of edges ranging from 1 000 000 to 10 000 000, measuring the running time of our methods. Fig. 4, shows that while `MinCov` and `Shapley` runtimes are below one second, SHC is three orders of magnitude slower than them. In summary, `MinCov` occupies a unique position in the algorithmic landscape, delivering near-optimal results with the linear-time efficiency required for large-scale networks.

## 8  Discussion and outlook

**Summary of contributions.** In this work, we introduced the CRITICALSET problem to quantify a particular notion of critical dependencies modeled in bipartite contributor-item networks. We proved that identifying the set of contributors whose removal causes the maximal collapse of content coverage is NP-hard and involves a supermodular objective function. This theoretical insight explains why standard forward-greedy strategies, popular in influence maximization, fail in this context. To overcome this, we derived a closed-form centrality based on the `ShapleyCov` value of a coalitional game and proposed `MinCov`, a linear-time peeling algorithm. Our experiments demonstrate that `ShapleyCov` and `MinCov` not only scale to massive graphs but also identify critical sets that are functionally invisible to the other competing methods.

**Impact of our study.** Our analysis contributes to the growing intersection of game theory and network science [33]. While the majority of recent work in this domain has focused on submodular maximization tasks, such as Influence Maximization [17], our work addresses the less-explored domain of supermodular maximization. By linking the CRITICALSET problem to the Shapley value, we provide a principled justification for our ranking method, distinguishing it from ad-hoc heuristics. This suggests a broader class of *criticality centralities* that could be developed for other network topologies beyond the bipartite case. Future work can extend our framework to dynamic and multilayer settings, investigating how contributors criticality evolves over time.

A direct application of our framework is the estimation of the criticality of open-source software, i.e. the minimum number of developers whose departure would stall a project (also known as Bus Factor). Current state-of-the-art methods largely rely on degree centrality or simple commit counts. Our results show that these metrics often underestimate criticality by ignoring the redundancy of knowledge distribution. CRITICALSET provides a formal, rigorous mathematical model for assessing this type of vulnerabilities. Future work should aim to validate the Shapley-derived centrality against real-world project abandonment data to establish it as a standard metric for repository health.

**Extensions to CRITICALSET: Weighted importance and soft thresholds.** Our current model assumes a worst-case scenario (AND logic): an item is lost only if all contributors depart, and all items are equally important. While appropriate for assessing total functional resilience, these definitions can be relaxed to cover a wider range of scenarios:

- Weighted Items: Real-world networks often contain items of varying value (e.g., a core software module vs. a documentation typo). A natural extension is to introduce weights $w_j$ for each item. If weights are uniform, the problem reduces to our current formulation.

---

[3] We carried out our experiments on a commodity laptop (Windows 11, 2.9 GHz CPU, 32 GB RAM).

- Soft Thresholds: The strict all-or-nothing condition could be generalized to a fractional threshold, where an item is considered lost if at least a given fraction of its contributors leave. This would bridge the gap between our supermodular model and linear degradation models.

**Theoretical outlook.** While the general hardness of CRITICALSET suggests that constant-factor approximations are unlikely, our ongoing work focuses on identifying specific graph classes where tighter bounds are achievable. We are currently investigating parameterized complexity approaches for small budgets $k$, as well as hybridized methods that adaptively transition between greedy and peeling strategies. By leveraging redundancy depth ($\phi_I, \phi_C$) and contributor uniqueness ($\gamma_C$), we aim to develop a second family of algorithms that dynamically adjust their selection logic to the local structure of the network.

**Acknowledgments.** This research is partially supported by MUR under PNRR project PE0000013-FAIR, Spoke 9 - Green-aware AI – WP9.2 and PN RIC project ASVIN "Assistente Virtuale Intelligente di Negozio" (CUP B29J24000200005).

**Disclosure of Interests.** The authors have no competing interests.